\documentclass{article}
\usepackage{arxiv}
\usepackage[numbers,sort&compress]{natbib}
\usepackage[utf8]{inputenc}
\usepackage[T1]{fontenc}
\usepackage{hyperref}
\usepackage{url}
\usepackage{array}
\usepackage{booktabs}
\usepackage{amsmath, amssymb, amsfonts}
\usepackage{microtype}
\usepackage{graphicx}
\usepackage{tikz}
\usetikzlibrary{shapes, arrows, positioning, fit, backgrounds, calc, trees}
\usepackage{amsthm}
\newtheorem{proposition}{Proposition}

\title{Latent State Design for World Models under Sufficiency Constraints}

\author{
 Keon Woo Kim \\
 Om Labs \\
 \texttt{keon@omlabs.xyz}
}

\begin{document}
\maketitle

\begin{abstract}
A world model matters to an agent only through the state it constructs. That state must preserve some information, discard other information, and support some future function: prediction, control, planning, memory, grounding, or counterfactual reasoning. This paper treats world-model research as latent state design under sufficiency constraints.

We propose a functional taxonomy that groups methods by what their latent state is for, rather than by architecture or application domain: predictive embedding, recurrent belief state, object/causal structure, latent action interface, grounded planning interface, and memory substrate. These roles expose distinctions that architecture-based groupings hide, including the gap between predictive sufficiency and control sufficiency, and the gap between passive video prediction and counterfactual action modeling.

The taxonomy supports an evaluation framework that judges a model by the sufficiency constraint its latent state was built to satisfy. We compare methods along seven axes: representation, prediction, planning, controllability, causal/counterfactual support, memory, and uncertainty. We use the resulting matrix as a diagnostic for what a latent state preserves, discards, and enables.

The conclusion that follows is that an actionable world model is the one whose state construction matches the task, not the one that preserves the most information.
\end{abstract}

\section{Introduction}
\label{sec:intro}

The problem with \textit{world model} is that the term names a function, not an architecture. In model-based reinforcement learning, it often means a latent dynamics model used for imagined control. In self-supervised video learning, it means a predictive representation of future visual structure. In generative simulation, it means a playable environment model. In object-centric work, it means a simulator over entities and relations. In robotics and vision-language-action systems, it means a planning interface grounded in perception, language, geometry, and action \citep{Hafner2019DreamControlBehaviors, Hafner2023MasteringDiverseDomains, Wu2022DayDreamerPhysicalRobot, Assran2025JEPASelfSupervised, DeepMind2025Genie3, NVIDIA2025CosmosWFM}. These uses share a commitment to internal state, but they ask that state to do different things.

Each use asks the latent state to be sufficient for something different. A predictive model asks for a state sufficient for future representation prediction. A control model asks for value- and action-relevant distinctions. An object-centric or causal model asks for persistent entities, relations, and intervention structure. A latent-action model asks for controllable change when explicit action labels are absent. A grounded planning model asks for latent geometry that supports search, goals, or trajectory choice. A memory model asks for persistence under partial observability. We call these requirements \textit{sufficiency constraints}.

This paper organizes world-model research by the function of latent state. Architectures matter, but they do not determine that function. The same backbone can serve prediction, control, memory, or planning depending on its objective, update rule, grounding signal, and evaluation. Conversely, different architectures can pursue the same sufficiency constraint. The right comparison is functional: what is the latent state for?

The framework has three parts. Section~\ref{sec:def} defines latent world models and states three propositions about how their sufficiency constraints relate. Sections~\ref{sec:predictive}--\ref{sec:memory} review the six roles. Section~\ref{sec:eval} turns the taxonomy into a seven-axis evaluation matrix. Section~\ref{sec:agenda} converts the framework into a falsifier-equipped research program.

A consequence is that a latent state is not better because it preserves more information. It is better when it preserves what its role requires and discards what interferes with that role. Reconstruction-heavy, predictive, control-oriented, causal, and memory-augmented models all sit on the same compression spectrum at different points, not on a single quality scale.

Adjacent work in LLMs, VLMs, and VLAs uses latent computation for reasoning, grounding, and action selection \citep{Hao2024Coconut, Bai2026ReasoningVLAThinking, Li2026ValueActionImplicit, Li2026GroundedSemanticallyGeneralizable, Wang2026WAMActionModeling}. Such systems enter the scope of world modeling only when their latent state represents an external world evolving under action. Without this restriction the term loses content.

The scope of this paper is conceptual rather than exhaustive. The methods we discuss are drawn from model-based reinforcement learning, self-supervised video learning, robotics, autonomous driving, object-centric modeling, latent-action learning, memory-augmented simulation, and vision-language-action systems. A method appears here when its latent state plays an explicit role in prediction, control, planning, grounding, memory, or counterfactual reasoning. Static representation learners are out of scope unless their latent state is used for temporal world evolution or downstream planning, and multi-role systems are placed under their primary latent-state function.

\section{What is a latent world model?}
\label{sec:def}

A \textbf{latent world model} is a learned state-update system for predicting how an external world evolves. This distinguishes it from a static representation model, which encodes semantics without modeling change, and from a purely generative decoder, which produces plausible outputs without maintaining a stable task-relevant internal state. The latent state stands in for the model's account of the environment, not merely for a compressed observation.

The boundary is defined by three properties. \textbf{State}: the latent must summarize aspects of the world beyond the current observation. \textbf{Temporal evolution}: it must change in a way that predicts or constrains future latent states or observations. \textbf{Action relevance}: under agent action, it must reflect the consequences of those actions. The definition is broader than a model-based RL definition and narrower than a purely generative one. It covers recurrent state-space models, predictive embeddings used to support planning, object-centric and causal models, grounded planning models, and memory-augmented systems where persistent latent state preserves world consistency over long horizons.

\subsection{Latent world models as state abstraction}
\label{sec:def-abstraction}

The boundary can be written as state abstraction under partial observability. An agent does not directly observe the full world state. It receives a history
\[
h_t = (o_1, a_1, o_2, a_2, \ldots, o_t),
\]
and learns a latent state abstraction
\[
z_t = \phi(h_t).
\]
A latent world model then learns how this state evolves under action:
\[
p_\theta(z_{t+1} \mid z_t, a_t).
\]

This formalism gives a common language for comparing what different models ask their latent states to preserve. The sufficiency tests differ by role: prediction asks whether $z_t$ supports future-observation forecasts; control asks whether it preserves value- and policy-relevant distinctions; object-centric modeling asks whether it factorizes the world into persistent entities; memory asks whether $z_t$ or its associated store preserves hidden state over long horizons.

Predictive sufficiency and control sufficiency differ. A latent state can support accurate future prediction while failing to organize states by reachability, value, or action consequences. Conversely, a compact control-sufficient state can discard details needed for photorealistic reconstruction. For latent state design, the relevant criterion is therefore which sufficiency the compression satisfies: prediction, control, structure, action abstraction, memory, or goal-conditioned planning \citep{Grimm2020ValueEquivalencePrinciple, Ferns2004Bisimulation, Castro2020DeepMDP, InformationBottleneck}.

The design problem can be read as a choice of \textbf{compression target}: each sufficiency constraint specifies what information the latent must retain and, by implication, what it should discard. From this view, the field's architectural variety reflects different choices along a compression spectrum, ranging from observation-faithful pixel reconstruction to value-equivalent or causally factored compression \citep{InformationBottleneck, Grimm2020ValueEquivalencePrinciple}. Reading the literature this way clarifies a recurring empirical pattern: control-oriented systems such as MuZero, EfficientZero, and TD-MPC2 move away from pixel-faithful reconstruction toward task-relevant compression \citep{Schrittwieser2020MuZero, Ye2021EfficientZero, Hansen2023TDMPC2}. Reconstruction can help when geometry or grounding is needed, but it becomes a liability when limited capacity is spent preserving details irrelevant to action.

Counterfactual support adds another requirement. If a world model is to answer what would happen under a different action, then its latent dynamics must be intervention-sensitive: changing $a_t$ should change the predicted future in a way that reflects possible consequences rather than passive correlation. Object-centric, causal, and action-conditioned latent models are therefore central to actionable world modeling \citep{Nam2026CausalJEPAObject, Ferraro2023FOCUSObjectCentric, Vakhitov2026ObjectCentricMeet, Gao2025AdaWorldAdaptableActions}, and sit near the most selective end of the compression spectrum.

\begin{table}[h]
\centering
\caption{Representative methods by compression-spectrum position. The year column shows how influential lines occupy different compression targets, from observation-faithful reconstruction to task-relevant abstraction.}
\label{tab:spectrum-methods}
\renewcommand{\arraystretch}{1.2}
\resizebox{\textwidth}{!}{%
\begin{tabular}{>{\raggedright\arraybackslash}p{3.0cm}>{\raggedright\arraybackslash}p{3.5cm}>{\raggedright\arraybackslash}p{2.0cm}>{\raggedright\arraybackslash}p{6.0cm}}
\toprule
Position & Training objective & Year range & Representative methods \\
\midrule
Reconstruction-heavy & Pixel reconstruction or generative modeling & 2018--2024 & World Models \citep{Ha2018WorldModels}, SimPLe \citep{Kaiser2020SimPLe} \\
Token compression & Discrete-token prediction & 2022--2025 & IRIS \citep{Micheli2022IRIS}, GAIA-1 \citep{Hu2023GAIA1}, GAIA-2 \citep{Russell2025GAIA2} \\
Representation prediction & Embedding-space prediction & 2023--2026 & I-JEPA \citep{Assran2023IJEPA}, V-JEPA 2 \citep{Assran2025JEPASelfSupervised}, V-JEPA 2.1 \citep{Labadia2026JEPAUnlockingDense}, LeWorldModel \citep{Maes2026LeWorldModelStableEnd} \\
Reward / value-shaped & Reward and policy-relevant supervision & 2019--2021 & TPC \citep{Nguyen2021TemporalPredictiveCoding}, value-aligned latent planning \citep{Havens2019StateSpacesPlanning} \\
Value-equivalent & Bellman-relevant statistics only & 2020--2023 & MuZero \citep{Schrittwieser2020MuZero}, EfficientZero \citep{Ye2021EfficientZero}, TD-MPC2 \citep{Hansen2023TDMPC2} \\
Causal / counterfactual & Intervention-sensitive structural variables & 2026 & Causal-JEPA \citep{Nam2026CausalJEPAObject}, CausalVAE-WM \citep{Ding2026CausalVAEWM} \\
\bottomrule
\end{tabular}}
\end{table}

Table~\ref{tab:spectrum-methods} maps design targets along this spectrum. Physical-reasoning probes \citep{Zhang2025MorpheusBenchmarkingPhysical, Liu2025CanSimulatorsReason} reinforce the axis by separating visual fidelity from physical and causal correctness.

\subsection{Relationships among sufficiency constraints}
\label{sec:def-sufficiency}

The six roles are descriptive. The sufficiency constraints behind them have formal relationships. Three propositions organize those relationships.

\begin{proposition}[Belief-state subsumption]
\label{prop:1}
Exact belief-state sufficiency subsumes prediction and control under standard POMDP assumptions. If $z_t$ is an exact belief state $b_t = p(s_t \mid h_t)$ and the transition, observation, and reward models are known, then $z_t$ is sufficient for predicting future observation distributions and for computing optimal values for any reward defined on the hidden state. This restates a standard result of POMDP theory \citep{Kaelbling1998POMDPs} and is the ideal case that recurrent latent world models approximate.
\end{proposition}

\begin{proposition}[Predictive vs.\ control sufficiency]
\label{prop:2}
Predictive sufficiency does not imply control sufficiency. A latent state can preserve information needed to predict future observations while discarding distinctions that affect reward or action choice. Formally, equality of predictive distributions over observations does not guarantee equality of $Q^*(h_t,a)$ unless the reward and action-relevant transition distinctions are also preserved. This follows from value-equivalence and bisimulation theory \citep{Grimm2020ValueEquivalencePrinciple, Ferns2004Bisimulation, Castro2020DeepMDP} and is why representation quality and planning quality must be evaluated separately.
\end{proposition}

\begin{proposition}[Passive vs.\ counterfactual sufficiency]
\label{prop:3}
Passive predictive sufficiency does not identify counterfactual dynamics without additional assumptions or interventions. A model trained only to predict observed transitions can learn correlations among events without learning how the system would evolve under alternative actions. This is the standard non-identifiability result from causal inference \citep{StructuralCausalModels}: counterfactual world modeling therefore requires action-conditioned data, causal assumptions, object/intervention structure, or grounding signals that constrain the latent dynamics.
\end{proposition}

Table~\ref{tab:proposition-failures} turns the propositions into evaluation failure modes. Each proposition identifies a common measurement error, the method families where the error appears, the benchmark that exposes it, and the latent-state design response.

\begin{table}[h]
\centering
\caption{Operational reading of the three sufficiency propositions.}
\label{tab:proposition-failures}
\renewcommand{\arraystretch}{1.2}
\resizebox{\textwidth}{!}{%
\begin{tabular}{>{\raggedright\arraybackslash}p{2.5cm}>{\raggedright\arraybackslash}p{3.5cm}>{\raggedright\arraybackslash}p{3.4cm}>{\raggedright\arraybackslash}p{3.6cm}>{\raggedright\arraybackslash}p{3.7cm}}
\toprule
Proposition & Evaluation mistake & Exposed in & Benchmark test & Latent design response \\
\midrule
Belief-state subsumption & Treating current-frame semantics or short-horizon prediction as evidence of state estimation & Predictive embeddings and VLA encoders without explicit memory & Occlusion, revisitation, delayed-cue, and loop-closure tasks & Recurrent filtering, retrieval, spatial memory, and state updates trained as belief approximation \\
Predictive vs.\ control sufficiency & Treating representation probes or video prediction as planning evidence & JEPA-style predictors and generative simulators evaluated outside closed-loop control & Matched frozen/adapted control, latent reachability, sparse-reward planning, and sim-to-real transfer & Action-conditioned dynamics, value-equivalent objectives, bisimulation losses, and planner-aware latent geometry \\
Passive vs.\ counterfactual sufficiency & Treating passive continuation likelihood as intervention evidence & Passive-video latent-action models and monolithic future predictors & Held-out action substitutions, object interventions, delayed causal effects, and counterfactual scene queries & Interventional data, object factorization, causal variables, and action-grounded latent transitions \\
\bottomrule
\end{tabular}}
\end{table}

The propositions connect the taxonomy to POMDPs, value equivalence, bisimulation, and causal modeling. They also map onto the compression spectrum: Proposition~\ref{prop:1} gives the belief-state target, while Propositions~\ref{prop:2} and~\ref{prop:3} explain why value-equivalent and causally factored compression require targeted evaluation rather than generic representation scoring.

\begin{figure}[h]
\centering
\resizebox{\textwidth}{!}{%
\begin{tikzpicture}[
 font=\scriptsize,
 axis/.style={->, >=stealth, very thick},
 tick/.style={thick},
 poslabel/.style={align=center, anchor=south, font=\scriptsize\bfseries, text width=2.4cm},
 ex/.style={align=center, anchor=north, font=\scriptsize\itshape, text width=2.4cm}
]
\draw[axis] (-0.4,0) -- (15.4,0);
\foreach \x/\name/\examples in {%
 1.0/{Reconstruction-\\heavy}/{World Models\\SimPLe},
 3.6/{Token\\compression}/{IRIS\\GAIA-1},
 6.2/{Representation\\prediction}/{I-JEPA\\V-JEPA 2\\LeWorldModel},
 8.8/{Reward /\\value-shaped}/{TD-MPC\\TPC\\value-aligned RSSM},
 11.4/{Value-\\equivalent}/{MuZero\\EfficientZero\\TD-MPC2},
 14.0/{Causal /\\counterfactual}/{Causal-JEPA\\CausalVAE-WM}}{
 \draw[tick] (\x,0.12) -- (\x,-0.12);
 \node[poslabel] at (\x,0.18) {\name};
 \node[ex] at (\x,-0.18) {\examples};
}
\node[anchor=north east, font=\scriptsize\itshape] at (-0.2,-0.05) {observation-faithful};
\node[anchor=north west, font=\scriptsize\itshape] at (15.2,-0.05) {task-relevant};
\end{tikzpicture}}
\caption{The compression spectrum of latent world model designs. Methods are ordered from observation-faithful reconstruction toward task-relevant compression, where each step trades richer sensory detail for a stronger sufficiency guarantee.}
\label{fig:compression-spectrum}
\end{figure}
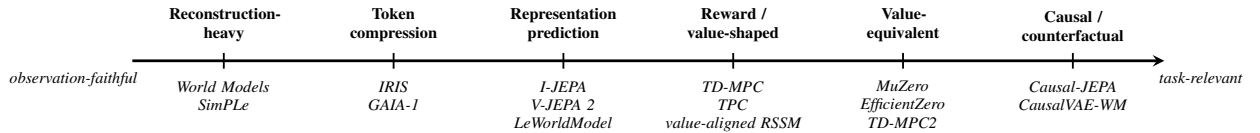

This boundary excludes many latent-space systems. Recent latent reasoning work in language and multimodal models shows that hidden representations can support reasoning, planning-like computation, or structured feature control. These systems qualify as world models only when they represent and evolve a state of the external world. The boundary is therefore not a stylistic preference but a scope decision: it preserves the explanatory content of the term \textit{world model}.

A method's qualification as a world model is measured by how strongly it satisfies the requirements of latent state, temporal evolution, and action-conditioned counterfactual support. Some papers emphasize representation while modeling action weakly. Others optimize explicitly for control but use less semantically transparent latent states. Still others prioritize memory, object structure, or grounding. The rest of the paper treats these differences as the main axes of the latent-state design problem.

This definition also clarifies why world models should be discussed separately from adjacent latent-space research in LLMs, VLMs, and VLAs. Those adjacent lines show that hidden spaces can support reasoning, grounding, and action-like interfaces, but they do not always require stable models of how external states evolve. The distinguishing burden of a world model is that its latent state must stand in for a changing world, not only a changing computation.

\section{A taxonomy of latent roles}
\label{sec:taxonomy}

The taxonomy has six roles: predictive abstraction, belief state, structured simulator state, action interface, planning geometry, and memory substrate. The roles overlap but are not interchangeable. Consistent with the compression-target framing of \S\ref{sec:def-abstraction}, each role specifies what the latent retains and what it discards.

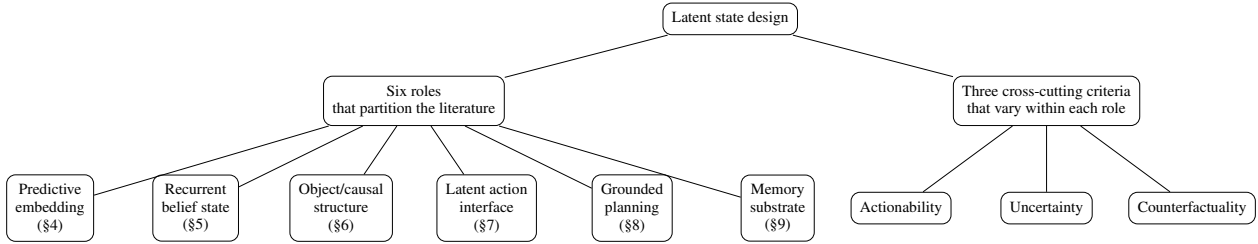
\begin{figure}[h]
\centering
\resizebox{\textwidth}{!}{%
\begin{tikzpicture}[
 level 1/.style={level distance=1.3cm, sibling distance=10cm},
 level 2/.style={level distance=1.7cm, sibling distance=2.3cm},
 every node/.style={draw, rectangle, rounded corners, align=center, font=\scriptsize, inner sep=4pt}
]
\node {Latent state design}
 child {node {Six roles\\that partition the literature}
 child {node {Predictive\\embedding\\(\S\ref{sec:predictive})}}
 child {node {Recurrent\\belief state\\(\S\ref{sec:dreamer})}}
 child {node {Object/causal\\structure\\(\S\ref{sec:object})}}
 child {node {Latent action\\interface\\(\S\ref{sec:latent-action})}}
 child {node {Grounded\\planning\\(\S\ref{sec:grounding})}}
 child {node {Memory\\substrate\\(\S\ref{sec:memory})}}
 }
 child {node {Three cross-cutting criteria\\that vary within each role}
 child {node {Actionability}}
 child {node {Uncertainty}}
 child {node {Counterfactuality}}
 };
\end{tikzpicture}}
\caption{The taxonomy of latent state design. Six roles partition the world-model literature by what the latent is primarily for; three cross-cutting criteria qualify how well a latent satisfies its role across all six.}
\label{fig:taxonomy}
\end{figure}

Each role corresponds to a different functional question a latent world state must answer: what will happen, what should be believed under partial observability, what entities persist, what changes are controllable, what futures are reachable, and what must be remembered. Architecture, modality, and domain cut across these roles rather than replacing them.

\subsection{Predictive embeddings}
\label{sec:taxonomy-predictive}

The first role treats latent state as a \textbf{predictive abstraction}. The goal is not to reconstruct every detail of the current observation, but to retain only the aspects of the world that matter for predicting future structure. This role is most visible in the JEPA family and related predictive representation learning systems \citep{Assran2025JEPASelfSupervised, Maes2026LeWorldModelStableEnd}. Here the latent is expected to suppress nuisance variation while preserving semantics, invariances, and dynamics-relevant content. The latent is judged on whether it represents the aspects of the world that make future states intelligible, not on whether it can regenerate the image.

This formulation challenges reconstruction-heavy views of representation learning, but predictive accuracy does not automatically imply planning readiness. A predictive embedding may preserve semantics without organizing the latent state space in a way that supports search, control, or value-sensitive decision-making. This gap motivates later branches of the taxonomy.

\subsection{Recurrent belief and latent state-space models}
\label{sec:taxonomy-belief}

A second design treats latent state as a \textbf{belief state} that evolves through time under action. The Dreamer family of model-based reinforcement learning systems makes this role canonical \citep{Hafner2019DreamControlBehaviors, Hafner2023MasteringDiverseDomains, Wu2022DayDreamerPhysicalRobot}. The latent state summarizes task-relevant information under partial observability, supports imagined rollouts, and provides a basis for policy learning or trajectory optimization. Unlike purely predictive embeddings, these states are explicitly optimized for control utility.

The design issue in this role is abstraction under action. A good latent state must preserve exactly the information that matters for decision-making while discarding visually irrelevant detail. The control-theoretic view of latent state is most explicit here, and the literature provides evidence that world models can improve sample efficiency and long-horizon planning.

\subsection{Object-centric and structured world state}
\label{sec:taxonomy-object}

A third design represents the scene as a \textbf{structured decomposition of the world} rather than as a single latent vector. Instead of representing the scene as a single monolithic vector, the model maintains object slots, particles, graph-structured entities, or other factorized state variables that correspond to persistent components of the environment. Such factorization is most useful in manipulation, physical reasoning, and compositional interaction \citep{Ferraro2023FOCUSObjectCentric, Wang2025DynBuildingStructured, Nam2026CausalJEPAObject, Vakhitov2026ObjectCentricMeet, Daniel2026ParticleSelfSupervised}.

The motivation is that many real-world tasks depend on persistent entities and their relations. Object-centric latent states promise better compositionality, clearer causal pathways, and more structured search. They may also provide a bridge between predictive latent learning and explicit planning, because a factorized world state can be easier to intervene on, manipulate, and reason about than a dense undifferentiated vector.

\subsection{Latent action interfaces}
\label{sec:taxonomy-action}

When fully labeled control signals are unavailable, latent state can serve instead as an \textbf{action abstraction layer}. Rather than assuming access to fully labeled control signals, the model learns an intermediate latent action space from passive video, demonstrations, or weakly structured interaction. The role connects observation-heavy data sources to embodied control \citep{Mendonca2023StructuredHumanVideos, Gao2025AdaWorldAdaptableActions, Klepach2025ObjectCentricAction, Wang2026FactoredAction}. Predictive world states represent what happens; latent actions represent what an agent can make happen.

The challenge is no longer only to model world evolution given known actions, but to discover an action-relevant factorization of change itself. Latent action models connect passive perception, affordance learning, and controllable planning.

\subsection{Grounded planning interfaces}
\label{sec:taxonomy-planning}

A fifth design treats latent state as a \textbf{planning interface} whose geometry is aligned with goals, values, language instructions, or physically meaningful structure. Here the question is not only whether the world state can be predicted, but whether it can be searched. Latent distances may be shaped by reachability, values, semantics, or trajectory feasibility. Recent grounded world models for robotics, autonomous driving, and VLA systems pursue this role \citep{Destrade2025ValueGuidedAction, Zhang2026HierarchicalPlanning, Li2026GroundedSemanticallyGeneralizable, Li2026ValueActionImplicit, Wang2026WAMActionModeling}.

The role exposes a weakness in otherwise strong representation models: semantically useful representations are not automatically planning-friendly. A latent planning interface must often satisfy much stricter requirements than a latent predictor. It must preserve distinctions that matter for action selection, be stable under rollout, and align with whatever supervisory signal defines good decisions.

\subsection{Memory-augmented world state}
\label{sec:taxonomy-memory}

Finally, latent state can serve as a \textbf{persistent memory substrate}. The goal is not merely to summarize the recent past, but to preserve world identity across revisitation, long horizons, partial observability, and camera or viewpoint shifts. This role becomes unavoidable once world models are asked to maintain consistency over long interaction sequences rather than short prediction windows.

Memory-augmented latent states can take several forms: recurrent compression, token memories, retrieval-based memory banks, or explicit spatial memory. What unifies them is that the latent state must be persistent enough to track a world rather than a frame sequence. This role appears in interactive video world models and embodied settings where loop closure and revisitation are necessary \citep{Wu2025VideoLongTerm, Xiao2025WorldMemLongTerm, Hong2025RELICInteractiveVideo, RecallToImagine}.

\subsection{Cross-cutting criteria: actionability, uncertainty, and counterfactuality}
\label{sec:taxonomy-criteria}

The six roles describe what latent state is \textit{used for}; three further criteria describe \textit{how well} a latent supports its role and apply across all six. \textbf{Actionability} asks whether latent state supports decision-making, control, or policy improvement. \textbf{Uncertainty} asks whether ambiguity and multi-modal futures are represented rather than collapsed away. \textbf{Counterfactuality} asks whether the model supports intervention-sensitive reasoning about what would happen under alternative actions. A predictive embedding or a memory representation may rate high or low on any of the three. The same role/criterion distinction reappears as the column structure of the matrix in \S\ref{sec:eval-matrix}.

\subsection{Roles versus architectures}
\label{sec:taxonomy-arch}

The six roles are often entangled within a single system, but conflating them obscures what has actually been achieved. A predictive embedding can be excellent at semantic abstraction while weak as a planning interface; a recurrent latent state-space model can excel in control while remaining hard to interpret; a memory-rich latent state can preserve identity over time without improving controllability. These are different compression targets, not partial successes against a single objective. Each latent role should be judged by whether it enables the function it was designed to serve, which is the basis of the evaluation framework of Section~\ref{sec:eval}.

\section{Predictive latent world models}
\label{sec:predictive}

The first major family in the taxonomy treats latent state primarily as a \textbf{predictive abstraction}. The idea is that a world model need not reconstruct every detail of the sensory stream. Instead, it should preserve the aspects of the world that make future states predictable and intelligible. This view is most closely associated with LeCun's JEPA program and its extensions to images, video, and embodied planning \citep{Assran2025JEPASelfSupervised, Labadia2026JEPAUnlockingDense, Maes2026LeWorldModelStableEnd}.

The JEPA family is one instantiation of a broader design space for predictive abstraction. Within this space, four axes vary across systems: the loss formulation, ranging over representation-space distance, contrastive, and masked-token reconstruction; the encoder/predictor asymmetry, joint or siamese; the masking strategy, with block, causal, or multi-block variants; and target encoder updates, by EMA, gradient stop, or learnable mechanisms. The methods cited in this section instantiate different combinations of these choices, but share the design commitment to predicting in a learned representation space rather than at the pixel level.

\subsection{The JEPA thesis: predict representations, not pixels}
\label{sec:predictive-jepa}

The JEPA line begins from a critique of reconstruction-heavy learning. Pixel-level prediction forces a model to allocate capacity to high-frequency details, texture, lighting, and other nuisance variation that may be irrelevant for reasoning or action. A joint-embedding predictive architecture instead encodes context and target observations into representation space and predicts the target representation from the context representation. In the notation introduced earlier, a context $x$ and target $y$ are encoded as
\[
z_x = f_\theta(x), \quad z_y = f_\theta(y),
\]
and a predictor estimates
\[
\hat{z}_y = g_\theta(z_x, c),
\]
with a representation-space loss
\[
\mathcal{L} = d(\hat{z}_y, z_y).
\]

The shift is not only architectural; it changes what the model is asked to preserve. Instead of all information needed to reconstruct the observation, the latent is trained to preserve information needed to predict another latent, discarding what cannot be predicted in embedding space.

JEPA's contribution is a clean separation between \textbf{world-relevant structure} and raw sensory detail: a state that is compact but still predictive of future-relevant information.

\subsection{From I-JEPA to video predictive representations}
\label{sec:predictive-video}

I-JEPA instantiates this idea for images by predicting missing or target image representations rather than reconstructing missing pixels. While image-level representation learning is not yet a full world model in the dynamic sense used here, it establishes the core principle: prediction can be moved into embedding space. This principle becomes more world-model-like when extended from static images to video.

V-JEPA 2 extends predictive latent learning into temporal visual data, making the latent responsible for representing structure across time \citep{Assran2025JEPASelfSupervised}. World modeling requires more than object recognition or static semantic abstraction; it requires some representation of temporal continuity, event structure, and change. In V-JEPA-style video models, the target is no longer only a missing image region but future or masked video structure in representation space. The video setting brings the JEPA program closer to the state-evolution requirement of world models.

V-JEPA 2 explicitly connects self-supervised video representations to understanding, prediction, and planning \citep{Assran2025JEPASelfSupervised}. In our taxonomy, this places V-JEPA 2 at a boundary: it is primarily a predictive latent model, but it begins to test whether predictive latents can support planning-relevant downstream behavior. By Proposition~\ref{prop:2}, representation-space prediction can improve future understanding without proving that the latent preserves reward- or reachability-relevant distinctions.

\subsection{Dense grounding and the move toward actionability}
\label{sec:predictive-dense}

A limitation of global or coarse latent representations is that they may support semantic understanding without preserving the dense spatial information needed for manipulation, navigation, or embodied control. V-JEPA 2.1 addresses this pressure by emphasizing dense visual features in video self-supervised learning \citep{Labadia2026JEPAUnlockingDense}. Physical action often depends on local geometry, object boundaries, affordances, and spatial correspondence. A latent state that is semantically meaningful but spatially coarse may be insufficient for action.

Dense predictive features are a step toward grounding. They do not by themselves solve planning or control, but they make predictive representations more usable for downstream embodied tasks. A broader pattern is visible here: predictive abstraction often has to be supplemented by grounding signals before it becomes actionable. Dense features, language alignment, value shaping, and geometry can all be understood as ways of making a latent state more usable for planning.

\subsection{LeWorldModel and stable predictive world modeling from pixels}
\label{sec:predictive-leworldmodel}

LeWorldModel continues the JEPA-inspired line by emphasizing stable end-to-end joint-embedding predictive world modeling from pixels \citep{Maes2026LeWorldModelStableEnd}. The work pushes predictive latent learning toward world-model use, asking whether a compact and stable predictive latent model can support planning without relying on massive generative simulation or complex multi-loss training.

Such efforts run counter to the world-foundation-model scaling narrative. Instead of assuming that larger video generators will become better simulators, LeWorldModel-style work investigates whether a smaller latent predictive system can be made stable, efficient, and planning-ready. It moves from representation sufficiency toward planning sufficiency, while retaining the JEPA commitment to representation-space prediction.

\subsection{What predictive latents give us, and what they do not}
\label{sec:predictive-summary}

Predictive latent world models provide a strong answer to one part of the world-model problem: how to learn abstractions that are less tied to raw sensory reconstruction and more tied to future-relevant structure. They are especially compelling as a response to the inefficiency of pixel-level prediction and as a route to semantic compression. They also provide a theoretical bridge to energy-based and self-supervised representation learning, where the goal is to shape a latent space without requiring dense labels or rewards.

However, predictive abstraction is not the same as actionability. Proposition~\ref{prop:2} gives the reason: predictive sufficiency does not imply control sufficiency. A latent state can be excellent for predicting future embeddings while still being poorly organized for search, reachability, or value-sensitive action selection. For example, two states may be semantically similar but require very different actions to reach a goal; conversely, two visually distinct states may be equivalent for control \citep{Grimm2020ValueEquivalencePrinciple, Ferns2004Bisimulation, Castro2021MICo}.

As an empirical matter, the JEPA line is strongest as a representation and transfer story. I-JEPA, V-JEPA 2, and V-JEPA 2.1 provide evidence that representation-space prediction yields semantic and dense visual features, while value-guided and hierarchical planning variants establish routes from those features toward action. The current evidence does not establish that JEPA-style predictive latents dominate closed-loop control benchmarks; it supports the narrower claim that predictive abstraction is a strong pretraining objective whose actionability depends on additional grounding, value, geometry, or planning supervision.

The open question for the JEPA-style world-model line is therefore:

\begin{quote}
When does prediction in representation space produce a latent state that is not only semantically meaningful, but also actionable?
\end{quote}

Recent work points to four answers: dense grounding \citep{Labadia2026JEPAUnlockingDense}, explicit planning objectives \citep{Destrade2025ValueGuidedAction}, hierarchical latent planning \citep{Zhang2026HierarchicalPlanning}, and stable end-to-end world-model training \citep{Maes2026LeWorldModelStableEnd}. The future of predictive latent world models depends less on representation learning alone and more on how predictive latents are connected to action, value, geometry, and memory.

\section{Latent imagination for control}
\label{sec:dreamer}

The second major family treats latent state primarily as a \textbf{substrate for action}. This is the model-based reinforcement learning line exemplified by PlaNet, the Dreamer family, and value-equivalent learners such as MuZero \citep{Ha2018WorldModels, Hafner2019PlaNet, Hafner2019DreamControlBehaviors, Hafner2023MasteringDiverseDomains, Wu2022DayDreamerPhysicalRobot, Schrittwieser2020MuZero, Ye2021EfficientZero}. Where JEPA-style systems ask what representation should be predicted, Dreamer-style systems ask what latent state lets an agent imagine consequences and improve behavior. This line foregrounds recurrent belief state and control sufficiency. Read along the compression spectrum, it sits further to the right than JEPA: not only are pixels discarded, but so is any structure not needed to reproduce optimal value.

\subsection{Latent imagination as a control loop}
\label{sec:dreamer-loop}

In Dreamer-style systems, the world model learns a latent dynamics model from interaction data. Observations are encoded into latent state, the transition model predicts how latent state changes under action, and imagined rollouts are used to train policies or value functions. Abstractly, the model learns a transition of the form
\[
p_\theta(z_{t+1} \mid z_t, a_t),
\]
and uses simulated latent trajectories
\[
z_t, a_t, z_{t+1}, a_{t+1}, \ldots
\]
as a training signal for decision-making. The latent is evaluated not by whether it predicts future observations but by whether it supports behavior.

The Dreamer family illustrates an \textbf{action-first} latent world model directly: the latent state succeeds if it preserves reward-, value-, and transition-relevant distinctions. A representation that clusters visually similar observations may still be poor for control if those observations imply different optimal actions.

\subsection{Control sufficiency and value relevance}
\label{sec:dreamer-control}

The theoretical grounding for this distinction comes from control sufficiency, value equivalence, and bisimulation. A latent state should preserve distinctions that affect the optimal value function or action-value function:
\begin{equation}
V^*(h_t) \approx V^*(z_t),
\end{equation}
or
\begin{equation}
Q^*(h_t,a) \approx Q^*(z_t,a).
\end{equation}

Value-equivalence theory makes this point explicit: for planning, a learned model does not need to preserve every detail of the environment, but it must preserve the consequences of actions that matter for value estimation and policy improvement \citep{Grimm2020ValueEquivalencePrinciple}. MuZero and EfficientZero implement this principle by learning latent dynamics whose only supervision is reward and policy consistency, demonstrating that a model can succeed at control while discarding observation-level detail \citep{Schrittwieser2020MuZero, Ye2021EfficientZero}. This is the control side of Proposition~\ref{prop:2}: a predictive latent and an actionable latent cannot be evaluated by the same probe, because the first compresses information needed to forecast and the second compresses information needed to decide.

Dreamer-style models test this distinction by training policies inside the learned latent dynamics. If imagined rollouts produce policy gradients or value targets that improve behavior, then the latent state has captured behaviorally relevant structure. This does not mean the latent is semantically transparent or causally factorized, but it does mean the latent has passed a control-oriented test.

\subsection{From simulated control to physical robots}
\label{sec:dreamer-robots}

DreamerV3 and TD-MPC2 extend the latent-imagination line toward general model-based RL across diverse continuous-control domains \citep{Hafner2023MasteringDiverseDomains, Hansen2023TDMPC2}. They make latent imagination look less like a task-specific trick and more like a general control mechanism. The same latent-dynamics idea can support behavior learning across domains with different visual statistics and reward structures.

DayDreamer pushes this line into physical robot learning \citep{Wu2022DayDreamerPhysicalRobot}. Embodied settings expose whether latent imagination can tolerate noisy observations, partial observability, imperfect dynamics, and real-world data constraints. In robotics, the latent state must support not only prediction but robust action under uncertainty. DayDreamer therefore connects model-based RL world models to physical AI.

The robotics setting also clarifies the limits of purely predictive representations. For an agent manipulating objects, the relevant abstraction is not only what the scene looks like or what future frames may contain, but which actions change the scene in controllable ways. The latent must represent affordances, contact-relevant structure, and action consequences well enough to improve policy learning.

\subsection{Temporal predictive coding and planning-relevant abstraction}
\label{sec:dreamer-tpc}

A related line studies latent spaces explicitly for model-based planning, including temporal predictive coding and reward-prediction-shaped latent state spaces \citep{Nguyen2021TemporalPredictiveCoding, Havens2019StateSpacesPlanning}. These works tie the design of latent state to the downstream planning problem: a latent trained only for observation reconstruction may preserve irrelevant detail, whereas a reward- or temporally-abstract-prediction-trained latent better preserves planning-relevant structure. The same environment can admit multiple latent states depending on whether the objective is prediction, reconstruction, planning, or control.

The strength of the Dreamer line is that it gives a concrete operational criterion for world modeling: does the learned latent dynamics help an agent act? This is a stronger test than visual plausibility or representation probing, and it grounds world modeling in policy improvement, sample efficiency, and closed-loop performance. Dreamer, DreamerV3, and DayDreamer demonstrate that latent imagination can drive policy learning across control domains and transfer into real robot learning, giving this family a strong closed-loop evidence base. The limitations are equally clear: learned states may be hard to interpret; reward-shaped latents may fail to preserve information needed for transfer; imagined rollouts compound model errors over long horizons; and standard MBRL relies on action-labeled data while modern embodied AI wants to exploit passive video, language, and heterogeneous demonstrations. These limitations motivate the later families: object-centric and causal latents add structure, latent-action models link passive observation to controllable change, grounded planning models align latent geometry with goals, and memory-augmented models preserve state over longer horizons.

\section{Object-centric and causal latent structure}
\label{sec:object}

The third major family asks whether a world model should represent the environment as a monolithic latent vector or as a structured collection of entities, relations, and interaction factors. Object-centric and causal world models argue, implicitly or explicitly, that many physical tasks require latent states that preserve \textbf{what objects exist}, \textbf{how they persist}, and \textbf{how they interact} \citep{Ferraro2023FOCUSObjectCentric, Wang2025DynBuildingStructured, Nam2026CausalJEPAObject, Vakhitov2026ObjectCentricMeet, Daniel2026ParticleSelfSupervised}. This family foregrounds object/causal structure. Read along the compression spectrum, it represents the rightmost active frontier: the latent retains essentially nothing except the entities and structural equations that an interventional query needs.

\subsection{Why factorize latent state?}
\label{sec:object-why}

A monolithic latent state can be sufficient for some prediction and control problems, but physical interaction often has compositional structure. Objects persist through time, interact locally, and can be manipulated independently or jointly. If a latent state collapses all of this into a single undifferentiated vector, then it may still support prediction, but it becomes difficult to intervene on, inspect, or search over.

Object-centric models address this by factorizing latent state into components such as slots, particles, entities, or graph nodes \citep{Mosbach2024SOLD, Jeong2025OCWMLang, Xu2025FewShotOCWM}. Abstractly, instead of a single latent state $z_t$, the model represents the world as
\[
z_t = (z_t^1, z_t^2, \ldots, z_t^n),
\]
where each component corresponds to an object-like or entity-like factor. Dynamics can then be modeled through local or relational updates:
\[
z_{t+1}^i = F_i\bigl(z_t^i, \{z_t^j\}_{j \in N(i)}, a_t, u_t\bigr).
\]

This factorization is attractive because it aligns with the structure of many embodied tasks. Manipulation, navigation, and physical reasoning often require tracking persistent entities, predicting interactions, and distinguishing changes to one object from changes to another.

\subsection{Object-centric world models for manipulation and interaction}
\label{sec:object-manipulation}

FOCUS is a representative example of object-centric world modeling for robotic manipulation \citep{Ferraro2023FOCUSObjectCentric}. It treats object structure not as an interpretability device but as a control-relevant representation. In manipulation, the agent often needs to reason about which object is being affected, what relation holds between objects, and how an action changes those relations.

Dyn-O and related structured world-model work extend this logic by building object-centric dynamics from pixels \citep{Wang2025DynBuildingStructured}. Object-centric structure matters most when it can be learned from raw perceptual input rather than provided symbolically. A latent world state that discovers object-like factors from pixels can potentially combine the scalability of representation learning with the compositionality of structured models.

Interactive object-centric reinforcement learning systems, including FIOC-WM, push this further by using object-centric world models for interaction and control \citep{Feng2025InteractiveObjectCentric}. These methods suggest that structured latent states can help agents reason about reusable interaction patterns rather than memorizing task-specific visual transitions.

\subsection{Causal latent structure}
\label{sec:object-causal}

Object-centricity is closely related to causal world modeling, but they are not identical. An object-centric model factorizes state; a causal model aims to capture how interventions change that state. For counterfactual reasoning, it is not enough to know that objects exist. The model must also represent how actions or interventions propagate through object relations.

Causal-JEPA makes this link explicit by introducing object-level latent interventions into the JEPA-style predictive framework \citep{Nam2026CausalJEPAObject}, and CausalVAE-style plug-ins push the same idea by attaching structural causal modules to encoder-transition backbones for counterfactual prediction \citep{Ding2026CausalVAEWM}. Causal-JEPA asks whether predictive latent representations can be made more intervention-sensitive by structuring the latent space around object-level changes. This is a natural extension of the predictive-latent line: if JEPA predicts representations, Causal-JEPA asks whether those representations can encode causal factors of variation rather than only statistical regularities.

For counterfactual world modeling, Proposition~\ref{prop:3} explains why passive prediction is insufficient: a predictor can learn that two events tend to co-occur without identifying how the system would change under intervention. An actionable world model should support questions of the form: what would happen if this object were moved, this action were taken, or this relation were changed? Object-centric causal latents are one path toward making such questions meaningful.

\subsection{Structured search and planning}
\label{sec:object-search}

Structured latent states can also make planning more tractable. Object-Centric World Models Meet Monte Carlo Tree Search uses object-centric world states in combination with search \citep{Vakhitov2026ObjectCentricMeet}. Factorization makes search more efficient and interpretable than search over dense continuous vectors. If the world is represented as interacting components, then a planner can exploit locality, modularity, and object-specific transitions.

Latent Particle World Models push a related idea through particle-like stochastic object-centric dynamics \citep{Daniel2026ParticleSelfSupervised}. The stochastic aspect matters because real environments are uncertain and partially observed. A structured latent state that also represents uncertainty is closer to what an embodied planner needs: not just a prediction of what will happen, but a distribution over what could happen to persistent entities.

Object-centric and causal latent models address several weaknesses of monolithic latent spaces: compositionality, structured search, intervention semantics, and a route toward counterfactual reasoning. They align naturally with physical interaction, where objects and relations are often the units over which humans and robots reason. In practice, the record is strongest where task structure really is entity-like: manipulation, object interaction, and search over compositional scene states, exemplified by FOCUS, Dyn-O, FIOC-WM, object-centric MCTS, and Latent Particle World Models. The evidence is weaker as a universal recipe: object discovery from raw pixels can be brittle, slot or entity decompositions may fail in cluttered scenes, deformable objects, or fluids, and not every control abstraction corresponds neatly to an object. Object-centric structure helps most when the environment's causal units match the factorization, and it still needs value, reachability, or task-goal shaping before it becomes planning-ready.

\section{Latent actions and passive-video-to-control bridges}
\label{sec:latent-action}

The fourth family focuses on a problem that becomes unavoidable when world models are trained from large observational datasets: videos show what happened, but they often do not specify what action caused it. For embodied agents, this creates a gap between passive prediction and controllable intervention. Latent-action models address this gap by learning internal variables that represent action-like or controllable changes, even when explicit action labels are unavailable or incomplete \citep{Mendonca2023StructuredHumanVideos, Gao2025AdaWorldAdaptableActions, Klepach2025ObjectCentricAction, Wang2026FactoredAction, Yang2026ChainThinkingMotion, Ye2024LAPO, Garrido2026LAMWild, Wang2025CoLAWorld}.

\subsection{Why latent actions matter}
\label{sec:latent-action-why}

A standard action-conditioned world model assumes access to pairs $(z_t, a_t, z_{t+1})$. This is natural in reinforcement learning and robotics datasets, but it is limiting for web-scale video, human demonstrations, and heterogeneous embodied data. Much of the available data contains observations of change without robot-native action labels. If world models are to exploit such data, then they need a way to infer what kind of controllable transformation occurred.

Latent-action models introduce an internal variable, say $\alpha_t$, that explains transition structure:
\[
p_\theta(z_{t+1} \mid z_t, \alpha_t).
\]
Here $\alpha_t$ is not a physical motor command by default. It represents a manipulation primitive, motion factor, object-level change, or abstract affordance. The goal is to learn a space of changes that can later be related to embodied actions.

Latent actions thus link predictive world models and control-oriented world models. Predictive models learn what changes; action-conditioned models learn what happens when an agent acts; latent-action models try to infer the missing middle: what change was implicitly enacted.

\subsection{Human-video world models}
\label{sec:latent-action-human}

Structured World Models from Human Videos provides a reference instance of this idea \citep{Mendonca2023StructuredHumanVideos}. Human videos contain rich affordance and interaction data such as objects moved, tools used, containers opened, and surfaces touched, but no robot-native action commands. The challenge is to extract action structure that transfers from observed human interaction to robot control: evidence about what changes are possible, which objects afford which transformations, and how interactions unfold over time. None of this is robot control, but it makes passive video a source of action-relevant supervision rather than only visual prediction.

\subsection{Adaptable and object-centric latent actions}
\label{sec:latent-action-adaptable}

AdaWorld explicitly studies adaptable world models with latent actions \citep{Gao2025AdaWorldAdaptableActions}. It treats latent actions as a mechanism for adapting world models across settings where explicit action spaces differ or are unavailable. A central issue in embodied AI is that action spaces are embodiment-specific, while observational data is embodiment-agnostic. A latent action space supplies an intermediate abstraction between the two.

Object-Centric Latent Action Learning adds another constraint: actions affect specific objects or relations rather than the whole scene uniformly \citep{Klepach2025ObjectCentricAction}. Combining latent actions with object-centric structure localizes controllable changes. Instead of learning a single global action factor, the model associates changes with object-level transitions.

Factored Latent Action World Models extend this logic to multi-entity settings \citep{Wang2026FactoredAction}. If a scene contains multiple entities, then a single monolithic latent action is insufficient. Different objects move, interact, or respond independently. Factored latent actions let the model represent multiple controllable changes within the same transition.

\subsection{Latent motion and VLA pretraining}
\label{sec:latent-action-vla}

Chain of World connects latent action or motion reasoning to vision-language-action pretraining \citep{Yang2026ChainThinkingMotion}. VLAs need to map perception and language to action, but direct action supervision is expensive and embodiment-specific. Latent motion chains represent action-relevant temporal structure before grounding it in a specific robot action space.

Latent-action world models and VLA systems converge here: both need an internal mapping from perceived change to executable action, and both treat latent actions as more action-relevant than pure visual embeddings while remaining short of motor commands.

The strength of latent-action models is strategic: action-labeled data is scarce, but passive video is abundant. Reliable latent action spaces let world models exploit large observational datasets while still supporting control. The empirical status is promising but less settled than the Dreamer-style record. Human-video and adaptable-world-model work demonstrates that passive observations contain action-relevant transition structure; object-centric and factored latent actions localize controllable change. The hard evidence that these latents transfer into robust closed-loop robot control is still thin. The identifiability problem is real: passive transition data admits many incompatible action explanations, so a learned latent action can encode visual motion without corresponding to anything an embodied agent can execute. Cross-embodiment transfer compounds this: a human hand, a robot gripper, and a mobile manipulator can produce similar visual changes but require different control policies. A latent action space must eventually be grounded in real actions, values, or goals; otherwise it remains an explanatory variable rather than an actionable interface.

\subsection{An action-grounding spectrum}
\label{sec:latent-action-spectrum}

The latent-action literature can be ordered along a single axis by how action enters the latent dynamics. The spectrum diagnoses whether a system is a world model, a policy, or something in between.

\begin{table}[h]
\centering
\caption{Action-grounding spectrum from R0 to R4: how action enters latent dynamics across world-model families.}
\label{tab:action-grounding}
\renewcommand{\arraystretch}{1.2}
\resizebox{\textwidth}{!}{%
\begin{tabular}{>{\raggedright\arraybackslash}p{0.8cm}>{\raggedright\arraybackslash}p{3.5cm}>{\raggedright\arraybackslash}p{6.3cm}>{\raggedright\arraybackslash}p{4.5cm}}
\toprule
Rung & Action relation & Examples & Capability \\
\midrule
R0 & no action & I-JEPA, V-JEPA, large-scale passive video generation & predictive abstraction or plausible futures \\
R1 & goal/text conditioning & Genie 3, language-conditioned simulators & weak controllability \\
R2 & inferred latent action & LAPO, AdaWorld, FLAM, LAM-Wild, CoLA-World & passive-video-to-control bridge \\
R3 & explicit action conditioning & Dreamer, MuZero, GAIA-2, OpenVLA & closed-loop control and planning \\
R4 & intervention-sensitive action & Causal-JEPA, CausalVAE-WM, CounterScene & counterfactual reasoning \\
\bottomrule
\end{tabular}}
\end{table}

Two observations follow from this ordering. Methods can be moved up the spectrum by adding action conditioning, value supervision, or causal structure, but this generally requires giving up data scale: higher rungs demand more constrained data sources. The rung at which a system sits also constrains what evaluations are meaningful: an R0 system cannot be tested for closed-loop control, and an R2 system cannot be tested for counterfactual queries without first establishing latent-action identifiability. Latent-action research is, in this view, the literature on climbing from R2 toward R3 and R4 without giving up the scale of passive video.

\section{Grounding, geometry, and planning interfaces}
\label{sec:grounding}

The fifth family focuses on what makes a latent state \textbf{planning-ready}. Prediction, control, object structure, and latent actions each address part of the world-model problem, but planning imposes another constraint: the latent space must be organized so that search, optimization, or trajectory inference is meaningful. This section reviews methods that ground latent state in value, hierarchy, language, geometry, or trajectory feasibility \citep{Destrade2025ValueGuidedAction, Zhang2026HierarchicalPlanning, Li2026GroundedSemanticallyGeneralizable, Li2026ValueActionImplicit, Wang2026WAMActionModeling, Zhou2026DriveDreamerPolicyGeometry}.

\subsection{From semantic similarity to planning geometry}
\label{sec:grounding-geometry}

A semantic representation groups states by meaning or appearance. A planning representation must group states by what can be done from them. The two are not equivalent: by Proposition~\ref{prop:2}, preserving predictive or semantic distinctions does not guarantee preservation of value, reachability, or action consequences. Two visually similar states may have very different action affordances; two visually distinct states may be equivalent for a task. This motivates a stronger requirement for planning-ready latent geometry.

In a planning-ready latent space, distances or directions correspond to quantities such as reachability, cost-to-go, value difference, or trajectory feasibility. For a goal state $z_g$, the geometry satisfies
\[
d(z_t,z_g) \approx \text{cost-to-go}(z_t,z_g),
\]
or more generally for the latent to support search over feasible trajectories. This does not mean all planning-ready latent spaces must be Euclidean or explicitly metric, but it does mean that arbitrary semantic embedding geometry is insufficient for planning.

\subsection{Value-guided JEPA and planning-aware predictive latents}
\label{sec:grounding-value}

Value-guided action planning with JEPA world models directly addresses the gap between predictive representation and action selection \citep{Destrade2025ValueGuidedAction}. It connects the JEPA-style predictive latent line to planning objectives. Rather than assuming that predictive embeddings are automatically useful for control, value-guided planning adds a signal that shapes or selects latent futures according to task value.

Value guidance offers a path from predictive sufficiency to control sufficiency: a representation-space predictor can learn future abstractions, but planning requires evaluating which futures are desirable or reachable.

\subsection{Hierarchical latent planning}
\label{sec:grounding-hierarchical}

Hierarchical Planning with Latent World Models addresses another planning difficulty: long horizons \citep{Zhang2026HierarchicalPlanning}. Flat latent rollouts can become unstable or expensive over many steps. Hierarchical planning decomposes the problem across temporal scales, allowing high-level latent plans to guide lower-level transitions.

This is theoretically natural. Many embodied tasks have hierarchical structure: navigate to an object, grasp it, move it, place it. A single-step transition model may not expose the right abstraction for such plans. A hierarchical latent world model can represent subgoals, temporally extended transitions, or multi-scale dynamics, making planning more tractable.

Hierarchy is a way of making grounded planning interfaces more effective rather than a separate latent role: it shapes latent space so long-horizon search becomes possible.

\subsection{Language-grounded planning}
\label{sec:grounding-language}

Grounded World Model for Semantically Generalizable Planning aligns latent world modeling with language-conditioned goals \citep{Li2026GroundedSemanticallyGeneralizable}. Embodied agents are increasingly expected to follow semantic instructions rather than optimize fixed reward functions. A latent planning space that is only value-grounded may be insufficient when goals are open-ended, compositional, or linguistically specified.

Language grounding changes the planning problem. The model must map instructions to latent goal conditions or scoring functions, and then use the world model to evaluate possible futures. In this setting, planning-readiness depends on whether the latent state preserves the distinctions that language can refer to: objects, relations, spatial configurations, affordances, and task-relevant changes.

The World-Value-Action Model similarly connects world modeling, value, and action in vision-language-action systems \citep{Li2026ValueActionImplicit}. It treats planning as latent inference over possible futures rather than as direct action generation. Treating planning as latent inference places VLA systems close to the world-model tradition: the model must internally represent what futures are possible and which are valuable.

\subsection{Geometry-grounded driving and world-action models}
\label{sec:grounding-driving}

Autonomous driving highlights the importance of geometry. In driving, planning depends on lane structure, object motion, spatial constraints, and trajectory feasibility. Earlier driving world models such as GAIA-1 and GAIA-2 show that scaled generative video models can act as driving simulators \citep{Hu2023GAIA1, Russell2025GAIA2}, and recent latent-state designs build on this lineage. Latent-WAM learns compact world-action tokens for end-to-end autonomous driving \citep{Wang2026WAMActionModeling}, while DriveDreamer-Policy emphasizes geometry-grounded world-action modeling for unified generation and planning \citep{Zhou2026DriveDreamerPolicyGeometry}.

In driving, planning-ready latent state cannot be purely semantic: it must preserve spatial layout, the motion of other agents, and feasible ego trajectories. Geometry then constrains the latent space toward physically meaningful futures.

Driving also reveals a broader evaluation issue: open-loop prediction quality may not predict closed-loop planning success. A latent world model for driving must be evaluated not only by whether it forecasts plausible futures, but whether it supports safe and effective decisions under interaction.

Grounded planning interfaces make latent world models more actionable by aligning latent state with value, reachability, language, geometry, or trajectory structure, directly addressing the main weakness of purely predictive latents: semantic quality does not guarantee searchability. This is where the field most clearly adjudicates against pure semantic representation: driving, robotics, and VLA systems need latents organized by feasibility, cost, and action consequences, as in value-guided JEPA, hierarchical latent planning, grounded world models, WVA, Latent-WAM, and DriveDreamer-Policy. Evidence is strongest in domain-specific benchmarks where the grounding signal is explicit. The tradeoff is that grounding can reduce generality: value-grounded latents are reward-specific, language-grounded latents depend on alignment quality, geometry-grounded latents may be domain-specific, and hierarchical latents depend on chosen temporal abstraction. Planning-ready design therefore involves a tradeoff between general representation and task-specific actionability.

\section{Memory and long-horizon latent consistency}
\label{sec:memory}

The sixth latent role is memory. Under partial observability, memory is part of state estimation rather than an optional module. A world model must often represent facts that are not visible in the current observation: object permanence, prior locations, hidden goals, map structure, delayed effects, or previously observed scene details. Recent work on memory-augmented world models makes this issue explicit \citep{RecallToImagine, Wu2025VideoLongTerm, Xiao2025WorldMemLongTerm, Hong2025RELICInteractiveVideo}.

Memory mechanisms in latent world models occupy a small design space along three axes. Storage substrate ranges over recurrent latent compression, transformer context, retrieval over an external bank, or spatial map. The update rule can be an online recurrent update, attention over context, or retrieval-and-write. The access pattern can be implicit consumption by the dynamics model or explicit query at planning time. The cited systems instantiate different points in this space, but share the design problem of approximating belief-state preservation under partial observability.

\subsection{Memory as belief-state approximation}
\label{sec:memory-belief}

In a fully observable Markov decision process, the current state is sufficient for prediction and control. In a partially observable setting, the agent only receives observations, so the sufficient statistic is a belief over hidden state:
\[
b_t = p(s_t \mid h_t).
\]
A learned latent state $z_t = \phi(h_t)$ can be viewed as an approximation to such a belief state. It compresses the history into a state representation that should preserve what matters for future prediction and action. By Proposition~\ref{prop:1}, exact belief state would subsume prediction and control; memory mechanisms approximate that sufficient statistic under capacity and data limits. Memory is therefore a theoretical necessity, not merely an engineering convenience.

Different world-model families implement memory differently. Recurrent state-space models compress history into recurrent latent state, an approach extended by early memory-augmented agents that combined external read-write memory with latent dynamics \citep{Wayne2018MERLIN}. Transformer world models may use token histories. Retrieval-augmented world models maintain external memory banks. Spatial-memory models preserve information about places, objects, or camera viewpoints. These mechanisms differ, but they address the same underlying problem: the current observation is not enough.

\subsection{Memory in control-oriented world models}
\label{sec:memory-control}

Recall to Imagine and related memory-focused world-model work study settings where an agent must remember information over time to act effectively \citep{RecallToImagine}. Short-horizon prediction benchmarks can hide memory failures. A model may predict the next few frames well while failing tasks that require recalling an object seen earlier, maintaining a map, or remembering a delayed cue.

\subsection{Long-term spatial memory and revisitation}
\label{sec:memory-spatial}

Video World Models with Long-term Spatial Memory addresses a key failure mode of generative and predictive video models: inconsistency over long horizons and revisitation \citep{Wu2025VideoLongTerm}. If a model generates or predicts a scene, leaves it, and later returns, then it should preserve the identity and layout of the world. Without memory, the model may hallucinate plausible but inconsistent details.

WorldMem similarly focuses on long-term consistent world simulation with memory \citep{Xiao2025WorldMemLongTerm}. The model must not only predict what comes next but maintain a record of what remains true when not currently visible.

A model that forgets hidden objects, previously visited rooms, or persistent spatial structure may still produce plausible observations, but it fails to model a stable world. Long-horizon memory is what distinguishes world modeling from short-horizon video prediction.

\subsection{Interactive worlds and long-horizon consistency}
\label{sec:memory-interactive}

RELIC studies interactive video world modeling with long-horizon memory \citep{Hong2025RELICInteractiveVideo}. The interactive setting raises the stakes because the model must preserve consistency while responding to user or agent actions. Memory is no longer only about making generated video look coherent; it is about maintaining a state that can support interaction over time.

Memory and counterfactuality therefore couple: different actions must update memory differently. A memory system that only retrieves past visual context is insufficient; it must integrate action-conditioned changes into persistent state.

Memory-augmented world models address persistence, the gap that short-horizon prediction hides. They track hidden state, preserve revisitation consistency, and maintain world identity over long interactions, which is essential for navigation, manipulation, interactive simulation, and any task with delayed consequences. The strongest evidence to date is on exactly the failures short-horizon video prediction overlooks: revisitation, object permanence, map consistency, and delayed information, as in WORLDMEM, RELIC, long-term spatial memory models, and Recall-to-Imagine. The benchmark ecosystem is still young, so the defensible claim is not that one memory architecture wins, but that memory should be tested as belief-state preservation under partial observability. Long memories can be expensive, retrieval can surface stale information, recurrent compression can forget detail, and spatial memory can be brittle under viewpoint or dynamic-scene changes.

\section{Design boundaries}
\label{sec:neighbors}

The role taxonomy in Section~\ref{sec:taxonomy} organizes the literature by what a latent state is \textit{for}. Two further classes of systems are worth treating separately because they are organized by \textit{architecture} rather than by role: generative world simulators, and sequence, state-space, and diffusion architectures that cut across multiple roles. Both are frequently called world models in current usage; both are most usefully read here as design boundaries rather than additional roles.

\subsection{Generative world simulators}
\label{sec:neighbors-generative}

Generative world simulators form a boundary case. Systems inspired by large-scale video generation, interactive environment generation, and physical-AI simulation platforms are often called world models because they produce plausible future observations or playable environments \citep{DeepMind2025Genie3, NVIDIA2025CosmosWFM, Yang2023UniSim}, but photorealistic or interactive generation is not identical to actionable latent state design.

A generative simulator can contain latent variables. What matters is whether those latents support planning, intervention, memory, and counterfactual control. A latent state that primarily supports visual synthesis can yield data generation or simulation while failing as a planning state. Action conditioning, value alignment, memory awareness, or geometric grounding moves it closer to the actionable notion. Along the compression spectrum, generative simulators sit furthest left: they retain what is needed to reconstruct plausible futures rather than discarding everything except what supports action.

Generalist agents trained inside such simulators push the boundary in the opposite direction by treating an LLM-style policy as the world-model interface. Generative simulators and generalist embodied agents are therefore neighboring paradigms rather than replacements for latent state analysis: they provide scale, visual richness, and interactive environments, while the latent-state perspective asks whether those systems also induce predictive, controllable, persistent, and counterfactual state.

\subsection{Sequence, state-space, and diffusion architectures}
\label{sec:neighbors-architectures}

Three architectural families recur across the role taxonomy and are easier to discuss together than to fold into any single role: trajectory-conditioned sequence models, recurrent state-space architectures, and diffusion-based generators. Each can implement a latent world model, but only under conditions worth making explicit.

\textbf{Trajectory-conditioned sequence models.} Decision-Transformer-style systems model behavior as sequences of states, actions, and returns, shifting some of the burden from explicit dynamics learning to conditional sequence modeling \citep{Chen2021DecisionTransformer}. These systems are not automatically world models under our definition: they can select actions by conditioning on desired returns without maintaining an explicit transition model. They become world-model-like only when their hidden state represents how the external state evolves across possible futures. A trajectory model qualifies as a world model only when its hidden state is sufficient for counterfactual action evaluation rather than behavior cloning on observed trajectories.

\textbf{State-space and Transformer architectures.} Mamba-style selective state-space models offer linear-time recurrent compression, while Transformer world models such as IRIS replace recurrence with attention \citep{Gu2023MambaSelectiveStateSpaces, Micheli2022IRIS}. Their appeal is efficient long-context filtering: the recurrent state update can compress extended histories without the quadratic cost of full attention. In latent-state terms, an SSM is attractive when its recurrent state approximates the belief statistic of Proposition~\ref{prop:1} better than a finite context window, especially under occlusion, delayed cues, or revisitation. The limitation is equally clear: efficient recurrence alone does not supply value equivalence, object factorization, or counterfactual intervention semantics. SSMs should therefore be evaluated as belief-state and memory mechanisms, not as automatic substitutes for action-conditioned world models.

\textbf{Diffusion policies and diffusion world models.} Diffusion-based policies and diffusion world models add a distributional alternative to single-rollout latent dynamics. Diffusion Policy represents visuomotor control as iterative denoising in action or trajectory space, which handles settings with several feasible ways to complete a task \citep{Chi2023DiffusionPolicy}. Diffusion world models similarly replace step-by-step deterministic rollout with generative modeling of possible futures, aiming to capture multi-modal trajectory distributions rather than a single predicted path \citep{Alonso2024DIAMOND}. For latent-state purposes, the diffusion variable counts as a planning state only when it conditions on goals or values, not when it merely samples plausible futures. When denoising is conditioned on goals, geometry, value, or interaction constraints, diffusion supplies a planning interface: it can represent multiple candidate futures, preserve uncertainty, and generate diverse feasible trajectories for downstream selection. When denoising only produces visually plausible futures, it sits closer to \S\ref{sec:neighbors-generative}'s generative-simulation boundary. Diffusion therefore contributes most to world modeling when its generative distribution is tied to closed-loop decision quality, not merely visual likelihood.

The common thread across all three architectural families is that their relation to world modeling is \textit{conditional on what the hidden state is required to support}, not on whether the architecture is recurrent, attentional, or generative. This is the same point made by the role taxonomy from a different angle: architecture does not fix sufficiency.

\section{Adjacent latent systems as boundary cases}
\label{sec:adjacent}

Latent reasoning in LLMs, VLMs, and VLAs belongs in this paper as a boundary section rather than as a seventh branch of the taxonomy. Language models use hidden representations for continuous or recurrent-depth reasoning \citep{Hao2024Coconut, Chen2026DeepThinkingTokens}; VLMs add visual grounding pressure \citep{Chen2024SpatialVLMEndowingSpatial}; VLA systems decode actions from multimodal latent state \citep{Kim2024OpenVLA}. These systems become world-model-like only when the latent state represents how an external environment evolves under action.

VLA systems sit closest to the boundary. Recent generalist robotic foundation models such as $\pi_{0.7}$ \citep{PhysicalIntelligence2026Pi07} combine subgoal image generation, video-history memory, and hierarchical policy structure. Other recent systems add explicit latent planning, trajectory prediction, or future-state modeling inside the policy \citep{Bai2026ReasoningVLAThinking, Li2026ValueActionImplicit, Wang2026WAMActionModeling}. Under the definition in Section~\ref{sec:def}, these systems enter the world-model category when their latent state carries action-conditioned consequences, not merely when it stores multimodal context.

The boundary is functional: a system is a world model when its latent is responsible for the consequences of action on the world. Latent computation alone is insufficient.

\section{Evaluation framework}
\label{sec:eval}

Evaluation must follow function. A model can generate plausible videos while failing as a planner, and a latent representation can support semantic probes while being poorly organized for control. Because latent roles differ, there is no single generic score for world modeling: evaluation must be tied to latent function \citep{Zhang2025MorpheusBenchmarkingPhysical}.

\subsection{Seven evaluation axes}
\label{sec:eval-axes}

The evaluation framework has seven axes. Four are role-aligned: representation, prediction, planning, and controllability. Three are cross-cutting criteria: causal/counterfactual, memory, and uncertainty. Representation and prediction quality are measured by probing, transfer, and forecast stability, which are necessary but not sufficient for action. Planning and controllability quality test whether the latent geometry supports search, reachability, and value-relevant action consequences; many semantically strong representations fail here. Causal and counterfactual quality asks whether the model represents interventions rather than passive correlation; memory consistency asks whether state survives occlusion, revisitation, and long horizons; uncertainty handling asks whether ambiguity and multi-modal futures are represented rather than collapsed away. Recent benchmarks make these axes operational: Morpheus, Gen-ViRe, and V-ReasonBench separate visual plausibility from physical and causal reasoning \citep{Zhang2025MorpheusBenchmarkingPhysical}; counterfactual and long-horizon memory suites directly test intervention-sensitive dynamics and persistent latent state; embodied and driving tasks expose the open-loop / closed-loop distinction. Each axis is named in the \S\ref{sec:eval-matrix} matrix and given a one-line evaluation question there.

\subsection{A functional evaluation matrix}
\label{sec:eval-matrix}

The evaluation framework is a matrix rather than a leaderboard. Rows are model families or papers; columns are latent functions and cross-cutting criteria:

\begin{table}[h]
\centering
\caption{Seven evaluation axes for actionable latent states.}
\label{tab:eval-axes}
\renewcommand{\arraystretch}{1.2}
\begin{tabular}{ll}
\toprule
Axis & Core question \\
\midrule
Representation & Does the latent state preserve meaningful structure? \\
Prediction & Does it forecast relevant futures? \\
Planning & Is the state space searchable for goals or actions? \\
Controllability & Do actions or latent actions change state in useful ways? \\
Causal/\hspace{0pt}counterfactual & Does the model support interventions and alternative futures? \\
Memory & Does it preserve hidden or persistent state over time? \\
Uncertainty & Does it represent ambiguity and multi-modal futures? \\
\bottomrule
\end{tabular}
\end{table}

The matrix compares model families without pretending they optimize the same thing.

To make the framework concrete, the table below applies it ordinally to ten representative methods spanning the six latent roles plus the generative-simulator boundary. The ratings are coarse and follow a single rubric:

\begin{itemize}
\item \textbf{P, primary:} the axis is named as a design target in the original work, the architecture or objective is \textit{constructed} to optimize it, and the paper reports evidence on at least one benchmark or task aligned with that axis.
\item \textbf{p, partial:} the axis is supported as a side-effect of the method's primary objective, or the original work reports incidental rather than targeted evidence on it; for instance, a control-oriented system that nonetheless yields usable predictive features.
\item \textbf{--, not targeted:} the axis is not part of the method's stated design intent, and no targeted evaluation is reported.
\end{itemize}

These ratings are \textit{functional}, not benchmark-derived: they describe what each method is \textit{designed} to satisfy. Patterns across rows and columns are the signal, not any single cell.

\begin{table}[h]
\centering
\caption{Functional evaluation matrix. P = primary design target, p = partial / side-effect, -- = not targeted.}
\label{tab:eval-matrix}
\renewcommand{\arraystretch}{1.2}
\resizebox{\textwidth}{!}{%
\begin{tabular}{lccccccc}
\toprule
Method / Role & Repr. & Pred. & Plan. & Ctrl. & Causal/CF & Memory & Uncert. \\
\midrule
V-JEPA 2 \citep{Assran2025JEPASelfSupervised} (predictive) & P & P & p & p & -- & p & p \\
DreamerV3 \citep{Hafner2023MasteringDiverseDomains} (recurrent belief) & p & P & P & P & -- & p & p \\
MuZero \citep{Schrittwieser2020MuZero} (value-equivalent control) & -- & P & P & P & -- & -- & -- \\
TD-MPC2 \citep{Hansen2023TDMPC2} (implicit latent control) & p & P & P & P & -- & p & p \\
FOCUS \citep{Ferraro2023FOCUSObjectCentric} (object-centric) & p & p & p & P & p & -- & -- \\
AdaWorld \citep{Gao2025AdaWorldAdaptableActions} (latent action) & p & p & p & p & -- & -- & -- \\
Grounded WM \citep{Li2026GroundedSemanticallyGeneralizable} (planning interface) & p & p & P & P & p & -- & -- \\
WORLDMEM \citep{Xiao2025WorldMemLongTerm} (memory) & p & P & p & -- & -- & P & p \\
DIAMOND \citep{Alonso2024DIAMOND} (diffusion WM) & P & P & p & P & -- & -- & P \\
Genie 3 \citep{DeepMind2025Genie3} (generative simulator) & P & P & -- & p & -- & p & p \\
\bottomrule
\end{tabular}}
\end{table}

Two patterns stand out. No row is filled across all axes: latent state design forces tradeoffs between representation, control, structure, memory, and uncertainty. And the sparse columns, Causal/CF and Uncertainty, are exactly the cross-cutting criteria from \S\ref{sec:taxonomy-criteria}. These columns identify the open frontier rather than diagnosing weakness in any single method.

Read in the compression-target language of \S\ref{sec:def-abstraction}, the matrix maps onto a single observation: methods further to the right of the spectrum, such as MuZero, TD-MPC2, and Causal-JEPA-style models, emphasize control or intervention-relevant structure while discarding representational and predictive richness that JEPA-style and generative-simulator methods retain. The spectrum makes that tradeoff legible.

\subsection{Companion matrix: preserve, discard, enable}
\label{sec:eval-companion}

The seven-axis matrix above asks, for each method, \textit{is the latent state designed to satisfy this criterion?} A complementary view asks the inverse: for each function, what should the latent \textbf{preserve}, what should it \textbf{discard}, and what should it \textbf{enable}? This phrasing makes explicit that discard is not a defect by default; a latent state must throw something away. It also aligns directly with the compression-target reading of \S\ref{sec:def-abstraction}.

\begin{table}[h]
\centering
\caption{Preserve/discard/enable companion matrix: what each latent function must retain, throw away, and unlock.}
\label{tab:companion-matrix}
\renewcommand{\arraystretch}{1.2}
\resizebox{\textwidth}{!}{%
\begin{tabular}{>{\raggedright\arraybackslash}p{2.6cm}>{\raggedright\arraybackslash}p{3.4cm}>{\raggedright\arraybackslash}p{3.0cm}>{\raggedright\arraybackslash}p{3.6cm}}
\toprule
Function & Preserve & Discard & Enable \\
\midrule
Prediction & future-relevant structure & pixel nuisance & stable forecasts \\
Control & value and action consequences & reward-irrelevant detail & policy improvement \\
Memory & hidden state over history & stale or irrelevant context & revisitation, delayed cues \\
Grounding & geometry, language referents, affordances & generic semantic similarity & goal-conditioned planning \\
Latent action & controllable change & embodiment-specific noise & passive-video-to-control transfer \\
Causal/\hspace{0pt}counterfactual & intervention factors & passive correlations & counterfactual queries \\
Uncertainty & distributional structure of futures & overconfident point estimates & safe rollout truncation \\
\bottomrule
\end{tabular}}
\end{table}

The companion matrix makes discard a first-class evaluation criterion alongside preservation, and reflects a recurring empirical signal: actionability tracks the fit of a method's discard pattern to its task more closely than the volume of information preserved.

\section{Research agenda}
\label{sec:agenda}

World-model benchmarks should not ask whether a model is generally good. They should ask \textbf{which sufficiency constraint its latent state satisfies}, and under what controlled conditions that claim holds. The five programs below test the framework by isolating one sufficiency constraint each, under matched data and compute, and naming what would falsify the claim about that constraint.

\subsection{Predictive sufficiency to action sufficiency}
\label{sec:agenda-predictive}

\textbf{Question.} When does a predictive latent become control-useful?

\textbf{Setup.} Pretrain on a shared passive video corpus under matched compute, then evaluate downstream control under both frozen and adapted protocols on standard benchmarks such as Atari, DeepMind Control Suite, and manipulation environments. Compare the predictive family of \S\ref{sec:predictive}, the control family of \S\ref{sec:dreamer}, and the latent-action family of \S\ref{sec:latent-action}.

\textbf{Metrics.} Sample efficiency, measured as interaction data needed to reach a fixed control threshold; planning success rate; and latent reachability under probe.

\textbf{Falsifier.} If frozen predictive latents match action-conditioned and value-shaped latents after equal adaptation budget, then predictive sufficiency is enough for this control setting.

\textbf{Predicted failure mode.} JEPA-style latents reach competitive representation-probe performance but require multiplicatively more interaction data to match Dreamer-style latents on closed-loop sample efficiency, especially under sparse reward.

\subsection{Counterfactual and interventional world modeling}
\label{sec:agenda-counterfactual}

The question this program addresses is whether a model learns intervention-sensitive dynamics or only plausible continuation statistics. Observed trajectories are shown to the model and outcomes are queried under altered actions, object interventions, hidden-state changes, and delayed effects, with explicit train/test interventional gaps; CounterScene-style benchmarks \citep{Jing2026CounterScene} provide a concrete instantiation. The relevant comparison is between monolithic predictive models, the object-centric and causal families of \S\ref{sec:object}, and the generative simulators of \S\ref{sec:neighbors-generative}, scored by counterfactual accuracy, intervention localization, and generalization to held-out interventions.

The case for explicit causal or object structure is overstated if monolithic predictive models match object-centric and causal models on out-of-distribution intervention queries after matched compute. Otherwise, the expected pattern is that monolithic predictive models succeed on action substitutions seen at training but degrade sharply on novel-object interventions and delayed causal effects, while object-centric and causal models degrade more gracefully because their factorization separates intervention loci.

\subsection{Memory as latent state under partial observability}
\label{sec:agenda-memory}

\textbf{Question.} When is memory genuine state estimation rather than longer context?

\textbf{Setup.} Environments with occlusion, revisitation, delayed cues, object permanence, changing hidden variables, and loop closure, at horizons that exceed every model's effective context window, so increasing context length is not a viable shortcut. Compare the memory mechanisms of \S\ref{sec:memory}: recurrent state-space, transformer context, retrieval, and spatial memory.

\textbf{Metrics.} Identity preservation, hidden-state recovery, revisitation accuracy, delayed-cue use, false-memory rate.

\textbf{Falsifier.} If a context-only transformer with sufficiently long context matches memory-augmented models on revisitation and delayed-cue tasks, then memory reduces to longer context and does not constitute a separate latent role.

\textbf{Predicted failure mode.} Long-context transformers close the gap on revisitation but fail delayed-cue tasks where information from one timestep must influence behavior thousands of steps later, because attention's effective context is structurally different from a recurrent belief approximator.

\subsection{Latent actions and embodiment transfer}
\label{sec:agenda-latent-action}

The program here asks whether latent actions represent controllable affordances or only visual motion clusters. The protocol pretrains on passive human or third-person video, then aligns inferred latent actions to robot actions in new embodiments under matched data and compute, holding out an embodiment and a task class. The latent-action models of \S\ref{sec:latent-action} are compared against inverse dynamics models trained directly on robot data and against VLA action decoders, with embodiment transfer rate, object-effect consistency, and action executability under held-out distractors as the relevant scores.

The case for latent-action pretraining from passive video weakens substantially if inverse dynamics models trained on robot data match the pretrained latent-action models after matched embodiment-alignment data. Otherwise, the expected pattern is selective: success on novel embodiments where the underlying object-effect structure is preserved, and underperformance on visually-similar / motor-distinct distractor tasks. The latter is the identifiability problem of Proposition~\ref{prop:3}.

\subsection{Closed-loop evaluation of generative simulators}
\label{sec:agenda-closed-loop}

\textbf{Question.} Do generative world simulators help agents act, or do they only generate plausible futures?

\textbf{Setup.} Place agents inside generated or predicted environments and measure transfer to real or high-fidelity environments, reporting visual-quality and closed-loop metrics jointly and treating divergence between them as the main scientific signal. Physical-reasoning benchmarks \citep{Zhang2025MorpheusBenchmarkingPhysical, Liu2025CanSimulatorsReason} provide established axes for the visual-vs-causal correctness gap. Compare the generative families of \S\ref{sec:neighbors-generative} against the latent dynamics models of \S\ref{sec:dreamer}.

\textbf{Metrics.} Policy improvement under sim-only training, sim-to-real transfer rate, exploitable-model-error rate, physical consistency.

\textbf{Falsifier.} If a high-visual-quality generator with no explicit dynamics model improves policy training as much as a latent dynamics model after matched compute, then the visual-vs-actionable distinction collapses.

\textbf{Predicted failure mode.} Agents trained inside the generator exploit specific generator artifacts, such as visual quality drift or physics violations near contact-rich edge cases, that do not transfer to real environments, and the divergence between visual-quality and closed-loop transfer scores becomes the main signal.

\section{Conclusion}
\label{sec:conclusion}

World-model research is latent state design under sufficiency constraints. Each method chooses what its learned state preserves, discards, and enables along a compression spectrum from observation-faithful reconstruction to causally factored compression. The unifying point across the six latent roles, the three sufficiency propositions, the evaluation matrix, and the research agenda is that a world model should be judged by which sufficiency constraint its latent satisfies.

An actionable latent state is defined by the fit between its discards and the task, not by how much observation detail it preserves.

\bibliographystyle{plainnat}
\bibliography{references}

\end{document}